\documentclass[11pt]{amsart}
\usepackage{geometry}
\usepackage[parfill]{parskip}    
\usepackage{graphicx}
\usepackage{amssymb}
\usepackage{epstopdf}
\usepackage{natbib}
\usepackage{hyperref}
\usepackage{amsmath}
\usepackage{stackengine}
\usepackage{xcolor}
\def\fauxat{\stackinset{c}{}{c}{}{\color{white}\scalebox{.01}{b0rked.org :metal: conform and obey. in theory, you are copying my email address. in practice, you are not.}}{@}}
  
\title{Cynical Selection of Language Model Training Data}
\thanks{Work unaffiliated with amazon.com}

\begin{document}

\maketitle
\begin{center}{amittai axelrod\\amittai\fauxat \hspace{0.1mm}alum.mit.edu}\end{center}

\begin{abstract}

The Moore-Lewis method of ``intelligent selection of language model training data" is very effective, cheap, efficient... and also has structural problems.
\begin{enumerate}
\item The method defines relevance by playing language models trained on the in-domain and the out-of-domain (or data pool) corpora against each other. This powerful idea -- which we set out to preserve -- treats the two corpora as the opposing ends of a single spectrum. This lack of nuance does not allow for the two corpora to be very similar. In the extreme case where the come from the same distribution, all of the sentences have a Moore-Lewis score of zero, so there is no resulting ranking.
\item The selected sentences are not guaranteed to be able to model the in-domain data, nor to even cover the in-domain data. They are simply well-liked by the in-domain model; this is necessary, but not sufficient.
\item There is no way to tell what is the optimal number of sentences to select, short of picking various thresholds and building the systems.
\end{enumerate}
We present ``cynical selection of training data'': a greedy, lazy, approximate, and generally efficient method of accomplishing the same goal. It has the following properties:
\begin{enumerate}
\item Is responsive to the extent to which two corpora differ.
\item Quickly reaches near-optimal vocabulary coverage.
\item Takes into account what has already been selected. 
\item Does not involve defining any kind of domain, nor any kind of classifier.
\item Has real units.
\item Knows approximately when to stop.
\end{enumerate}

\end{abstract}

\section{Scenario \nopunct \\}
We have a number of translation tasks, each defined as ``a flavor of data we want to be able to translate well''. Perhaps it's a specific client's data, or a kind of data such as ``customer support chat logs'', or just a system for a language arc (``Mexican Spanish to American English'). Machine Translation (MT) tasks often have one of these two problems:
\begin{enumerate}
\item Web-scale data, or too much data than can be used readily to train or run an MT system. We want to know what data we can {\em exclude} from training without sacrificing performance. 
\item Not enough parallel data to train an MT system. We want to know what data we can {\em add} to improve performance, and how much improvement we might expect. We can add data from known parallel data, or by paying to have monolingual data translated.
\end{enumerate}
We present a single method for handling both:
\begin{enumerate}
\item Given a too-large parallel corpus, we identify the best subset to use for training a system that is at least as good as the full system. This is the traditional data selection scenario, and we improve upon the standard Moore-Lewis cross-entropy difference method \citep{MooreLewis:2010}.
\item Given a representative monolingual corpus for a translation task, and an optional parallel corpus in the language pair, identify the monolingual sentences that should be manually translated and added to the parallel corpus in order to improve translation for the specific task.
\end{enumerate}
Translating monolingual data is expensive, and training on bilingual data is expensive. By ``expense'' we mean effort, computation, time, and dollars. As a rule, we want to spend as little as possible to get as much out of the available data as we can.

\section{Context}

\subsection{Language Model Size \nopunct \\}

\cite{SeymoreRosenfeld:1996} decided that pruning a model is better than incrementally growing it:
\begin{quotation}Using more training data, up to at least 25~-~30 million words initially, and then pruning it down is a better approach than just starting with a small amount of training data [...] Beyond 25 million words, the amount of training data does not have a noticeable effect.\end{quotation}

\cite{Stolcke:1998} showed how to produce an efficiently-sized language model (LM) by building an LM and then pruning entries in the model that do not help model the training set. This is determined by whether removing the entries changes the entropy of the training set, or not. 
They ``assume that the $n$-grams affect the relative entropy roughly independently, and compute [the change in entropy] due to each individual $n$-gram [...] then rank the $n$-grams by their effect on the model entropy'' and focus on those that affect the model entropy the least. The relative entropy score of the sentence actually {\em can} be decomposed into the sum of the word scores (see derivation by \cite{SethyGeorgiouN:2006}), so there is no need to assume.

\cite{SiivolaPellom:2005} approach from the other direction: incrementally growing an n-gram LM, adding entries that decrease the number of bits required to model the training data. They compute gains on the data coding length, along the lines of the Minimum Description Length principle which minimizes the size of the model plus the training data as encoded by the model \citep{Rissanen:1983}.
\cite{SethyGeorgiouN:2006} also take the growing approach, using an information-gain -based score. However, due to computational cost, they did not update the estimate after every new selection.

\subsection{Corpus Similarity \nopunct \\}

\cite{Kilgarriff:2001} posited: \begin{quotation} Corpus similarity is complex, and there is no absolute answer to ``is Corpus 1 more like Corpus 2 than Corpus 3?''. All there are are possible measures which serve particular purposes more or less well. \end{quotation}
Perhaps similarity cannot be defined absolutely, but the units of measurement can. Bits of information are ideal for this.\\

\subsection{Data Selection \nopunct \\}

Data selection -- the deliberate reduction in size of the training corpus -- is a roundabout way of achieving domain adaptation. Each of the training data sentences is scored by its relevance to the data to be modeled, and only the best-scoring sentences are kept. The standard approach for data selection is the same for both language modeling \citep{MooreLewis:2010} and machine translation \citep{AxelrodHeG:2011}. Called ``cross-entropy difference'', it preferentially selects data that is {\em like} the translation task and {\em unlike} the rest of the data pool.\footnote{ More complete explanations can be found in previous work: \cite{AxelrodHeG:2011} and \cite{Axelrod:2014}.}

\subsection{Cynically Quantifying Relevance \nopunct \\}

One major advantage to the above-mentioned model-growing methods is that there is no need to train a huge language model in order to figure out how to build a smaller one. An advantage of information-theoretic similarity measurements is that the focus is on quantifying the relationship between sentences and corpora, so the output can be used for a variety of scenarios and not just language modeling (or, by extension, machine translation). 

Here we propose a method to incrementally construct an efficiently-sized downstream model by incrementally constructing an efficiently-sized training corpus. We score data by how \underline{useful} it would be to add it to a corpus that is used to model a particular translation task. In particular, we score sentences by how much we'd learn if we added it to our existing data, right now. \cite{SethyGeorgiouN:2006} describe the general idea as ``an incremental greedy selection scheme based on relative entropy, which selects a sentence if adding it to the already selected set of sentences reduces the relative entropy with respect to the in-domain data distribution".

\section{Inputs}
The algorithm requires four inputs to be specified, twice as many as the Moore-Lewis cross-entropy difference method.
which only had two: the in-domain and the general-pool data files. We separate the idea of the vocabulary distributions of the kinds of language that we have (or want) from the actual corpora that we have (or want). These additional inputs allow the cynical selection method to be used in a wider range of applications.

\subsection{REPRESENTATIVE corpus distribution \nopunct \\}
A corpus that is representative of what we want to translate. For example: all the data we have for a task, or a sample of a data stream from the domain, or whatever. We will evaluate the selected data by its ability to model the \textsc{Representative} corpus (specifically: by computing the perplexity of the \textsc{Representative} corpus using a language model trained on the selected data).
We only use the \textsc{Representative} file to compute some vocabulary statistics (described later). This means a client with confidential data they can't share could just share the derived statistics and not the text itself.

\subsection{UNADAPTED corpus distribution \nopunct \\}
A corpus that reflects our currently-available data. This was called the ``data pool'' in prior work, but we've changed the name to make some distinctions clearer. We only need this corpus to compute some corpus statistics, and then compare the statistics of this corpus with that of the \textsc{Representative} corpus. Perhaps the \textsc{Unadapted} is generic data, perhaps it is just all the data that we have in the language of interest. By default the \textsc{Unadapted} corpus distribution would come from the set of sentences to select from (\textit{i.e.} the data pool is the unadapted corpus is the set of sentences to be scored).

The \textsc{Representative} corpus gives us the vocabulary distribution of the data we {\em want}, and \textsc{Unadapted} gives us the vocabulary distribution of the data we {\em have}. We need these two distributions in order to quantify how these two sets of language differ.

\subsection{SEED text file \nopunct \\}
Contains any headway that has been made towards assembling a corpus that meets the goal of being able to train a system to translate Representative-type data. It might be that the client has already sent off 1000 hand-picked sentences to be translated, or we have 100k sentences we are certain come from Spanish-speaking customers in Mexico or whatever the unifying theme of the \textsc{Representative} corpus is. The \textsc{Seed} is assumed to be empty by default.

\subsection{AVAILABLE text file \nopunct \\}
The set of candidate sentences we can choose {\em from}. We will score and rank these sentences according to their usefulness in modeling the \textsc{Representative} distribution. In most cases, the \textsc{Available} corpus is the same one that is used to compute the \textsc{Unadapted} corpus distribution.

However, if the goal is to pick monolingual sentences to send off for manual translation, then the \textsc{Available} corpus is probably the \textsc{Representative} corpus! In this case, the \textsc{Representative} corpus would be monolingual client data: things we want to know how to translate, but have no translations for. The \textsc{Unadapted} is probably our existing parallel data, and we want to know how much parallel \textsc{Representative} data we need to purchase and add to the \textsc{Unadapted} in order to build a system to translate the rest of the \textsc{Representative} data.

\subsection{JADED text file \nopunct \\}
This is the output file. It is a re-ranked subset of the \textsc{Available} corpus with a few special (cynical) properties. Sentences in the file are ordered (and scored) by the amount of useful information they contain about the \textsc{Representative} corpus. Specifically, we measure how much we can improve (decrease) the perplexity of the \textsc{Representative} corpus evaluated with a language model trained on the \textsc{Seed}+\textsc{Jaded} data.
Other practical properties:  if we read the file line by line, everything already read is more useful than everything we have not yet seen. Furthermore, the next line in the file is always the most informative sentence to add to what we have already seen.  
Every sentence in the file is ranked strictly according to its utility.

\section{Notation}
\begin{itemize}
\item The lowercase {\tt w} variable is for word tokens. \\
Any capital {\tt W} is some set of tokens in a corpus.
\item The lowercase {\tt v} variable is for word types (lexical / vocabulary items). \\
Any capital {\tt V} is a set of vocabulary items in a corpus.
\item The lowercase {\tt c} variable is for the count of a single word type. \\
Any capital {\tt C} variable is the count of some word type in a corpus.
\item $w_n$ is the set of tokens in the $n^{th}$ selected line. \\
$v_n$ is the set of word types in the $n^{th}$ selected line. \\
$c_n(v)$ is the count of word type $v$ in the $n^{th}$ line.
\item $W_n$ and $V_n$ are the total number of word tokens and word types, respectively, in the first $n$ selected lines. $C_n(v)$ is the count of word $v$ in the first $n$ selected lines. 
\item $W_{\textsc{repr}}$ and $V_{\textsc{repr}}$ are the total number of word tokens and word types, respectively, in the \textsc{representative} corpus. Guess what $W_{\textsc{unadapt}}$ and $V_{\textsc{unadapt}}$ are?
\end{itemize}

\section{Math}

\subsection{Cross-Entropy: What It Is \nopunct \\}
Let $Q$ be the probability distribution from a language model $LM$. The general definition of the cross-entropy $H$ of the \textsc{Repr} corpus using $Q$ is:
\begin{align}
H_{LM} (\textsc{repr}) &= - \sum_{v \in V_{\textsc{repr}}} \frac{C_{\textsc{repr}}(v)}{W_{\textsc{repr}}} \log Q(v)
\end{align}
$H$ represents how well the language model's training corpus can be used to model the \textsc{Representative} corpus.
\textit{Perplexity} is normally used to evaluate language models. The general definition is: $$ppl_{LM}(\textsc{repr}) = 2^{H (\textsc{repr})}$$
Everything that follows could be rewritten in terms of perplexity, but we won't.

The stupidest kind of language model we can build is an unsmoothed unigram LM (the maximum likelihood estimate LM). That kind of LM makes $Q$ be the empirical probability of each word in the corpus:
\begin{align}
H (\textsc{repr}) &= - \sum_{v \in V_{\textsc{repr}}} \frac{C_{\textsc{repr}}(v)}{W_{\textsc{repr}}} \log \frac{C(v)}{|W|}
\end{align}
$H$ is always positive, and smaller (lower entropy) is better. \\

We will be selecting sentences to incrementally grow a corpus. After selecting the $n^{th}$ line, the cross-entropy between the $n$-sentence corpus and the \textsc{Representative} one is:
\begin{align} \label{eqn:XEnt-n}
H_n (\textsc{repr}) &= - \sum_{v \in V_{\textsc{repr}}} \frac{C_{\textsc{repr}}(v)}{W_{\textsc{repr}}} \log \frac{C_n(v)}{W_n}
\end{align}
Call that $H_n$ for short. We will be computing $H_1, H_2, \ldots, H_{n-1}, H_{n}$ as we select lines.

\subsection{Cross-Entropy: What It Is Not \nopunct \\}
Entropy, cross-entropy, and relative entropy are easily confused in the literature. They are three different things. Compounding the situation, there are also alternate names: `Entropy' specifically means `Shannon entropy', and `Relative Entropy' means the Kullback-Leibler (KL) divergence. The relationship between the kinds of (Shannon) entropy regarding two probability distributions $P$ and $Q$ is:
\begin{align*}
\textrm{Entropy} & = \textrm{Cross Entropy} && - \textrm{Relative Entropy}\\
& = H (P,Q) && - D_{KL} (P||Q)\\
& = - \sum_x p(x) \log q(x) && - \sum_x p(x) \log \frac{p(x)}{q(x)}\\ 
& = - \sum_x p(x) \log q(x) && - \sum_x p(x) ( \log p(x) - \log q(x) )\\ 
& = - \sum_x p(x) \log q(x) && - \sum_x p(x) \log p(x) + \sum_x p(x) \log q(x)\\
H (P) & = - \sum_x p(x) \log p(x)
\end{align*}

An intuitive explanation\footnote{ Expanded from anonymous Stanford NLP notes on KL Divergence:\\
\url{https://nlp.stanford.edu/fsnlp/mathfound/fsnlp-slides-kl.pdf}}: The Shannon entropy of $P$ is the minimum number of bits required to perfectly encode all events drawn from distribution $P$. Cross entropy is the number of bits required to encode all events drawn from $P$, but using an imperfect code based on distribution $Q$. Relative entropy is the number of bits that are wasted by using $Q$ instead of the true distribution $P$. Then: Number of bits used - wasted bits = optimal number of bits.

\subsection{Defining Cross-Entropy Greedily\nopunct \\}

Suppose we have selected $n$ sentences, and have computed the current subset's score $H_n$. We want to know what is the next sentence we should select to achieve our goal of finding the subset of sentences that minimize $H$. After the $n+1^{th}$ step we wish to have the best (lowest) possible $H_{n+1}$ score. We cannot un-select a sentence, so $H_{n+1}$ can be defined as:\\
\begin{align} \label{eqn:delta-H}
H_{n+1} = H_n + \underset{n\rightarrow n+1}{\Delta H}
\end{align}

It is sufficient to find the $n+1^{th}$ sentence that minimizes\footnote{ Sign-switching is hard: $H$ ideally decreases over time, making $H_{n+1} < H_n$. Then $[H_{n+1} - H_n]$ is negative, and ideally as negative as possible. It could be less complicated to define $H_{n+1} = H_n - |\underset{n\rightarrow n+1}{\Delta H}|$ and then maximize the positive quantity $|\Delta H|$.} the negative quantity $\underset{n\rightarrow n+1}{\Delta H}$.

We will shortly derive the following:
\begin{align} \label{eqn:delta-H-alone}
\underset{n\rightarrow n+1}{\Delta H} = \underset{n\rightarrow n+1}{\textrm{Penalty}} + \underset{n\rightarrow n+1}{\textrm{Gain}}
\end{align}
so of course:
\begin{align} \label{eqn:inductive-H}
H_{n+1} = H_n + \underset{n\rightarrow n+1}{\textrm{Penalty}} + \underset{n\rightarrow n+1}{\textrm{Gain}}
\end{align}

The penalty term is positive, and the gain term is a negative number. Adding any new line to the corpus incurs an entropy-increasing penalty. If the candidate line is a good one to add to the corpus, then it adds useful information to the model and will lower the cross entropy. A sufficiently good candidate sentence will provide a gain that outweighs the penalty. We would like to minimize the (positive) penalty term, and also minimize the gain term (as negative as possible).

\subsection{Deriving the Greedy Cross-Entropy Delta\nopunct \\}
Expanding the definition of $\Delta H$ (Equation~\ref{eqn:delta-H}) using Equation~\ref{eqn:XEnt-n}:
\begin{align*}
\underset{n\rightarrow n+1}{\Delta H} &=  H_{n+1} - H_n\\
&= \left(- \sum_{v \in V_{\textsc{repr}}} \frac{C_{\textsc{repr}}(v)}{W_{\textsc{repr}}} \log \frac{C_{n+1}(v)}{W_{n+1}} \right)
  - \left( - \sum_{v \in V_{\textsc{repr}}} \frac{C_{\textsc{repr}}(v)}{W_{\textsc{repr}}} \log \frac{C_n(v)}{W_n} \right)\\
&= \left( \sum_{v \in V_{\textsc{repr}}} \frac{C_{\textsc{repr}}(v)}{W_{\textsc{repr}}} \log \frac{C_n(v)}{W_n} \right)
 - \left( \sum_{v \in V_{\textsc{repr}}} \frac{C_{\textsc{repr}}(v)}{W_{\textsc{repr}}} \log \frac{C_{n+1}(v)}{W_{n+1}} \right)\\
&= \sum_{v \in V_{\textsc{repr}}} \frac{C_{\textsc{repr}}(v)}{W_{\textsc{repr}}} 
\left( \log \frac{C_n(v)}{W_n} - \log \frac{C_{n+1}(v)}{W_{n+1}} \right)\\
&= \sum_{v \in V_{\textsc{repr}}} \frac{C_{\textsc{repr}}(v)}{W_{\textsc{repr}}} 
\left( \log \frac{\frac{C_n(v)}{W_n}}{\frac{C_{n+1}(v)}{W_{n+1}}} \right)\\
&= \sum_{v \in V_{\textsc{repr}}} \frac{C_{\textsc{repr}}(v)}{W_{\textsc{repr}}} 
\log \left( \frac{C_n(v)}{W_n} \cdot \frac{W_{n+1}}{C_{n+1}(v)} \right)\\
&= \sum_{v \in V_{\textsc{repr}}} \frac{C_{\textsc{repr}}(v)}{W_{\textsc{repr}}} 
\log \left( \frac{W_{n+1}}{W_n} \cdot \frac{C_n(v)}{C_{n+1}(v)} \right)\\
&= \sum_{v \in V_{\textsc{repr}}} \frac{C_{\textsc{repr}}(v)}{W_{\textsc{repr}}} 
\left( \log \frac{W_{n+1}}{W_n} + \log \frac{C_n(v)}{C_{n+1}(v)} \right)\\
&= \sum_{v \in V_{\textsc{repr}}} \frac{C_{\textsc{repr}}(v)}{W_{\textsc{repr}}} \log \frac{W_{n+1}}{W_n}
+ \sum_{v \in V_{\textsc{repr}}} \frac{C_{\textsc{repr}}(v)}{W_{\textsc{repr}}} \log \frac{C_n(v)}{C_{n+1}(v)} \\
&= \log \frac{W_{n+1}}{W_n} \left( \sum_{v \in V_{\textsc{repr}}} \frac{C_{\textsc{repr}}(v)}{W_{\textsc{repr}}}\right)
+ \sum_{v \in V_{\textsc{repr}}} \frac{C_{\textsc{repr}}(v)}{W_{\textsc{repr}}} \log \frac{C_n(v)}{C_{n+1}(v)} \\
&= \log \frac{W_{n+1}}{W_n} + \sum_{v \in V_{\textsc{repr}}} \frac{C_{\textsc{repr}}(v)}{W_{\textsc{repr}}} \log \frac{C_n(v)}{C_{n+1}(v)} \\
\underset{n\rightarrow n+1}{\Delta H} &= \underbrace{\log \frac{W_n + w_{n+1}}{W_n}}_{Penalty} + 
\underbrace{\sum_{v \in V_{\textsc{repr}}} \frac{C_{\textsc{repr}}(v)}{W_{\textsc{repr}}} \log 
\frac{C_n(v)}{C_n(v) + c_{n+1}(v)}}_{Gain}
\end{align*}

We have derived Equation \ref{eqn:delta-H-alone}, the greedy \underline{cross}-entropy delta. \cite{SethyGeorgiouN:2006:naacl} and (\citeyear{SethyGeorgiouN:2006}) derived instead the greedy \underline{relative}-entropy delta. Interestingly, both kinds of entropy have the same deltas because the extra $p(x)$ terms cancel out. Our Penalty term is the same as their $T1$ term, and our Gain term is their $T2$ term.\footnote{ Actually, our Gain $= -(T2)$, because we gather the negative sign inside the term and they do not.}

The penalty term biases the delta score towards selecting short sentences. Adding any new line to the corpus incurs an entropy-increasing penalty proportional to the number of tokens in the new line, regardless of what the words are. The penalty term is always positive, and decreases asymptotically to zero: adding one more sentence makes less and less of a difference as the selected corpus grows larger. A specific sentence's penalty decreases over time, too, for the same reason.
 
The gain term biases the delta score towards longer sentences that contain many words which are high in probability in the \textsc{Repr} distribution but do not appear many times in the sentences selected so far. There is an entropy-lowering benefit to adding information, so the gain of adding any sentence is always a negative number\footnote{ Negative or zero, so... non-positive?} that gets subtracted from $H_n$. Adding more data can be net good (the magnitude of the gain term is larger than the penalty, so $H$ decreases) or net bad (the magnitude of the gain is less than the penalty, increasing $H$). A sufficiently good candidate sentence will provide a gain that outweighs the penalty: it is adding useful information to the model, and will lower the cross-entropy.

The length biases of the penalty and the gain terms counteract each other, guarding the algorithm from the Moore-Lewis method's fixation on one-word sentences with a very common token, or long sentences full of OOVs.

We can plug Equation \ref{eqn:delta-H-alone} into Equation \ref{eqn:inductive-H} to get an iterative method for computing cross-entropy:
\begin{align*}
H_{n+1} &= H_n &&+ \underset{n\rightarrow n+1}{\textrm{Penalty}} &&+ \underset{n\rightarrow n+1}{\textrm{Gain}}\\
&= H_n &&+ \underbrace{\log \frac{W_n + w_{n+1}}{W_n}}_{Penalty} &&+ \underbrace{\sum_{v \in V_{\textsc{repr}}} \frac{C_{\textsc{repr}}(v)}{W_{\textsc{repr}}} \log 
\frac{C_n(v)}{C_n(v) + c_{n+1}(v)}}_{Gain}
\end{align*}

Again: the log terms force the penalty to be positive and the gain to be negative.

\subsection{Algorithmic Complexity of Greedy Cross-Entropy Delta Method \nopunct \\}

Suppose we have already selected the $n$ most useful sentences from \textsc{Available}, removing them from their original corpus and placing them into \textsc{Jaded}. The theoretically optimal algorithm picks sentence $n+1$ as follows:
\begin{enumerate}
\item Compute the $\underset{n\rightarrow n+1}{\Delta H}$ score for each sentence remaining in \textsc{Available}. This requires going word by word through the sentences in \textsc{Available} to compute the Penalty and Gain terms based on the current count of each word in the sentence.
\item Sort the sentences in \textsc{Available} by $\underset{n\rightarrow n+1}{\Delta H}$ score, and choose the sentence with the best (lowest) score to be sentence $n+1$. Remove it from \textsc{Available} and add it to \textsc{Jaded}.
\end{enumerate}

The steps are repeated for each subsequent sentence to be chosen.

The cost appears brutally prohibitive: $O(N)$ iterations, each requiring an update of $O(W_{\textsc{avail}})$ words and then a sort of $O(N \log N)$ lines. The number of words in the corpus is roughly linear with respect to the number of lines ($N = 15\textrm{-ish } W$), so the cost is $O(N^3 \log N)$.

We will not quite do that.

\section{Hulk Smash Math}
 
\subsection{Good-Enough Sentences \nopunct \\}
\label{sec:good-enough-sentences}
The Gain term for a particular sentence after selecting $n$ other ones is always:
$$\underset{n\rightarrow n+1}{\textrm{Gain}} = \sum_{v \in V_{\textsc{repr}}} \frac{C_{\textsc{repr}}(v)}{W_{\textsc{repr}}} \log 
\frac{C_n(v)}{C_n(v) + c_{n+1}(v)}$$
The total lexicon $V_{\textsc{repr}}$ can be divided into two parts: vocabulary items $v_{n+1}$ that do appear in the $(n+1)^{st}$ sentence, and those that do not ($v \notin v_{n+1} = V_{\textsc{repr}} \setminus v_{n+1})$ :
\begin{align*}
\underset{n\rightarrow n+1}{\textrm{Gain}}
&= \sum_{v \in V_{\textsc{repr}}} \frac{C_{\textsc{repr}}(v)}{W_{\textsc{repr}}} \log 
\frac{C_n(v)}{C_n(v) + c_{n+1}(v)}\\
&= \sum_{v \in v_{n+1}} \frac{C_{\textsc{repr}}(v)}{W_{\textsc{repr}}} \log 
\frac{C_n(v)}{C_n(v) + c_{n+1}(v)}
+ \sum_{v \notin v_{n+1}} \frac{C_{\textsc{repr}}(v)}{W_{\textsc{repr}}} \log 
\frac{C_n(v)}{C_n(v) + 0}\\
&= \sum_{v \in v_{n+1}} \frac{C_{\textsc{repr}}(v)}{W_{\textsc{repr}}} \log 
 \frac{C_n(v)}{C_n(v) + c_{n+1}(v)} + \sum_{v \notin v_{n+1}} \frac{C_{\textsc{repr}}(v)}{W_{\textsc{repr}}} \cdot 0 \\
 &= \sum_{v \in v_{n+1}} \frac{C_{\textsc{repr}}(v)}{W_{\textsc{repr}}} \log 
 \frac{C_n(v)}{C_n(v) + c_{n+1}(v)}\\
 &= \sum_{v \in v_{n+1}} \underset{n\rightarrow n+1}{\textrm{Gain}} (v)
\end{align*}
The words that do not appear in the sentence do not affect the $\Delta H$ score. The gain for the sentence is the sum of the gains for each word (type) in the sentence. The empirical probability $\frac{C_{\textsc{repr}}(v)}{W_{\textsc{repr}}}$ has the usual long-tail distribution, so we hypothesize that the sentence gain will often be dominated by only one or a few of the word (type) gain terms. Those dominant words will be ones that the \textsc{Jaded} corpus needs to see more times in order to accurately model the \textsc{repr} distribution.

In that case, it would always be useful to select a sentence that contains the word with the best word gain. The best-gain sentence that contains the best-gain word may not be the overall best-gain sentence, but it will still be pretty good. It does contain the single most useful word, so even if would ideally not choose to pick this sentence right now, we would probably want to pick it before too long.

The exact gain for a sentence from containing a specific word type depends on $c_{n+1}(v)$, the number of times the word type appears in the sentence. We hypothesize that most words will appear once per sentence. The obvious exception are the closed-class words, and they are few. As such, we can estimate the word gain as:
$$\underset{n\rightarrow n+1}{\textrm{Gain}} (v) = \frac{C_{\textsc{repr}}(v)}{W_{\textsc{repr}}} \log 
 \frac{C_n(v)}{C_n(v) + c_{n+1}(v)} \approx \frac{C_{\textsc{repr}}(v)}{W_{\textsc{repr}}} \log 
 \frac{C_n(v)}{C_n(v) + 1}$$
This estimate is an \textit{upper} bound on the true value of the word gain, because the gain term is negative and $\log \frac{A}{A+1} > \log \frac{A}{A+2}$. This means we are selecting the best word based on a lower bound estimate of the magnitude of its goodness, which is reasonable. The true word gain for any word is at least as good (more negative) as its estimate, so the best word is at least as good as we think it is. We now select sentence $n+1$ as follows:

\begin{enumerate}
\item Estimate the $\underset{n\rightarrow n+1}{\textrm{Gain}} (v) \approx \frac{C_{\textsc{repr}}(v)}{W_{\textsc{repr}}} \log \frac{C_n(v)}{C_n(v) + 1}$ for all word types $v \in V$.
\item Sort the word types in $V$ by their estimated gain, and note the best word $v'$
\item Let $\textsc{Avail}(v')$ be the set of lines still in \textsc{Available} which contain $v'$.\\
Compute the $\underset{n\rightarrow n+1}{\Delta H}$ score for each sentence in $\textsc{Avail}(v')$.\\
This requires going word by word through each of those sentences to compute the Penalty and Gain terms based on the current count of each word in the sentence.
\item Sort $\textsc{Avail}(v')$ by the $\underset{n\rightarrow n+1}{\Delta H}$ score, and choose the sentence with the best (lowest) score to be sentence $n+1$. Remove it from \textsc{Available} and add it to \textsc{Jaded}.
\end{enumerate}

This algorithm takes $O(N)$ iterations, each requiring:\\
an update of $V$ word gains and a sort of $O(V \log V)$ words to get the best word $v'$,\\
and then an update of $O(W_{\textsc{Avail}(v')})$ words and a sort of $O(|\textsc{Avail}(v')| \cdot \log |\textsc{Avail}(v'))|$ lines to get the best line containing $v'$.\\
The number of words in those lines $W_{\textsc{Avail}(v')}$ is again linear w.r.t. the number of lines, or $O(|\textsc{Avail}(v')|)$. The size of $|\textsc{Avail}(v')|$ can be estimated by the average number of lines that a specific word appears in. Empirically, it's a little less than $ \sqrt[3]{N} = N^{\frac{1}{3}}$, as well as smaller than both $\log^2 N$ and $\log^2 V$. We'll use the $N^{\frac{1}{3}}$ term for clarity.
This reduces the complexity to $O( N \left(V^2 \log V + N^{\frac{1}{3}} +  N^{\frac{1}{3}} \log (N^{\frac{1}{3}}) \right))$, which is mostly $O( N  V^2 \log V)$ because large-set vocabularies have over a million word types.

\subsection{Good-Enough Words \nopunct \\}

We have reduced the time complexity of the optimal algorithm to be heavily dependent on the size of the vocabulary rather than the number of lines in the training data. We now focus on reducing the size of the vocabulary itself, with the following intuition \citep{AxelrodVyasMC:2015}: \begin{quote}``Where the frequencies of words differ, the corpora differ. Where the frequencies do not differ, neither do the corpora. [...]
We use the ratio of the word's probabilities in the corpora to determine how much the two specific corpora differ with respect to a word. [...]. This can also be readily computed using unigram LMs trained on each of the corpora."\end{quote}

Consider the unigram frequency ratio  $\frac{P_{\textsc{repr}}(v)}{P_{\textsc{unadapt}}(v)}$ of an arbitrary word $v$. There are four cases of interest to us:
\begin{enumerate}
\item{$C_{\textsc{repr}}(v) < 3$ \underline{and} $C_{\textsc{unadapt}}(v) < 3$}:\\
Barely-seen words do not have reliable empirical statistics. A word appearing twice in one corpus and once in another is not necessarily twice as likely, nor is a singleton that does not appear in the other corpus infinitely more likely. The unigram ratio is therefore also not reliable. \\
The threshold of 3 is not important; it was picked because LM smoothing is often applied to terms appearing $< 3$ times. Choose a threshold $x$ based on the question: ``If I saw a word $x-1$ times, would I trust its probability?''. It does not need to be the same threshold for both corpora, either.

\item{$\frac{P_{\textsc{repr}}(v)}{P_{\textsc{unadapt}}(v)} << 1$}\\
When the ratio is considerably less than\footnote{ An order of magnitude less or so: $< \frac{1}{e}$, $< \frac{1}{2}$, or $< \frac{1}{10}$ depending on base. } 1, the word is strongly indicative of the \textsc{Unadapted} distribution. These words (by definition) appear relatively rarely in the \textsc{Repr} corpus, so cannot comprise a significant chunk of the cross-entropy $H$.
\item{$\frac{P_{\textsc{repr}}(v)}{P_{\textsc{unadapt}}(v)} \approx 1$}\\
These words appear roughly as often (within an order of magnitude) in one distribution as in the other. A sample of the \textsc{Avail} corpus can be expected to contain these words in a reasonable proportion.
\item{$\frac{P_{\textsc{repr}}(v)}{P_{\textsc{unadapt}}(v)} >> 1$}\\
When the ratio is considerably greater than\footnote{ An order of magnitude more or so: $> e$, $> 2$, or $> 10$ depending on base. } 1, the word is strongly indicative of the \textsc{Representative} distribution. These words are the most important ones: They comprise much of the probability mass of the \textsc{Repr} distribution (according to how much \textsc{Repr} and \textsc{Unadapt} diverge), and they are relatively rare in the \textsc{Avail} corpus.
\end{enumerate}

We assert that the words in the fourth category are most important, and that it suffices to use only them for selecting sentences. We collapse each of the first three categories to a single token, again following \cite{AxelrodVyasMC:2015} and \cite{Axelrod:2014}. Category 1 words become \texttt{`dubious'}, category 2 are \texttt{`bad'}, and category 3 are \texttt{`meh'}. The words in category 4 remain unchanged. We further differentiate words in \textsc{Repr} but not \textsc{Avail} as \texttt{`impossible'}, and words in \textsc{Avail}+\textsc{Unadapt} but not in \textsc{Repr} as \texttt{`useless'}. This replacement strategy consolidates probability mass for negative and neutral events, letting the distribution focus on words whose presence is meaningful and relevant.

Empirically, the reduces the number of words in the vocabulary to around 10,000 word types: words $V_{+REPR}$ with heavy bias towards the \textsc{repr} distribution. The complexity calculation in Section \ref{sec:good-enough-sentences} becomes 

\begin{enumerate}
\item Estimate the $\underset{n\rightarrow n+1}{\textrm{Gain}} (v) \approx \frac{C_{\textsc{repr}}(v)}{W_{\textsc{repr}}} \log \frac{C_n(v)}{C_n(v) + 1}$ for the  $\approx$10,000 word types in $V_{+REPR}$.
\item Sort the word types in $V_{+REPR}$ by their estimated gain, and note the best word $v'$
\item Let $\textsc{Avail}(v')$ be the set of lines still in \textsc{Available} that contain $v'$.\\
Compute the $\underset{n\rightarrow n+1}{\Delta H}$ score for each sentence in $\textsc{Avail}(v')$.\\
\item Sort $\textsc{Avail}(v')$ by $\underset{n\rightarrow n+1}{\Delta H}$ score, and choose the sentence with the best (lowest) score to be sentence $n+1$. Remove it from \textsc{Available} and add it to \textsc{Jaded}.
\end{enumerate}

This algorithm takes $O(N)$ iterations, each requiring:\\
an update and then sort of $V_{+REPR}$ word gains, but $O(10,000)$ is now constant: $O(C)$.\\
An update of $O(W_{\textsc{Avail}(v')})$ words and a sort of $O(|\textsc{Avail}(v')| \cdot \log |\textsc{Avail}(v'))|$ lines to get the best line containing $v'$.\\
The number of words in those lines $W_{\textsc{Avail}(v')}$ is again $< N^{\frac{1}{3}}$. 
The complexity is $$O( N \left(C + N^{\frac{1}{3}} + N^{\frac{1}{3}} \log N^{\frac{1}{3}} \right)) \approx O( N N^{\frac{1}{3}} \log N^{\frac{1}{3}}) = O( N^{\frac{4}{3}} \cdot \frac{1}{3} \log N) = O( N^{\frac{4}{3}} \log N) $$
This value compares very favorably with the original $O(N^3 \log N)$.

\subsection{Good-Enough Scoring \nopunct \\}
\label{sec:good-enough-scoring}

The bulk of the processing time in the algorithmic block is in the rescoring and re-sorting of the sentences that contain the best word for that particular timestep. We assert we do not need to score all the sentences every time.

The sentence $\Delta H$ score estimate term is the sum of the penalty and gain terms. While both the penalty and the gain decrease over time, their sum is not monotonic. The sentence gain estimate 
$\left( \sum_{v \in V_{\textsc{repr}}} \frac{C_{\textsc{repr}}(v)}{W_{\textsc{repr}}} \right)$ for a particular sentence always increases over time (the term becomes less negative, so the sentence has less of an effect on the entropy score). A sentence's current gain estimate is thus a lower bound on the sentence's gain estimate at some future iteration.

The sentence penalty $\left( \log \frac{W_n + w_{n+1}}{W_n} \right)$ for a particular sentence always decreases over time. However, if we compare the magnitude of the derivatives, the penalty term $\frac{d}{dn} Penalty$ appears to be smaller than $\frac{d}{dn} Gain$ as $n \to \infty$.\footnote{  The following argument might hold, but we are not sure:\\
Using standard identities, $\frac{d}{dx} \left( \log \frac{x + k}{x}\right) = \frac{-k}{x(x+k)}$ and $\frac{d}{dy} \left( \log \frac{y}{y+i}\right) = \frac{i}{y(y+i)}$ , so $\frac{d}{dn} Penalty \approx \frac{-k}{W_n(W_n + k)}$ and $\frac{d}{dn} Gain \approx \sum_v \frac{i}{C_n(v)(C_n(v) + i)}$
In general, $k$ (avg number of words per sentence) $> i$ (avg number of occurrences of a word within a sentence), and $W_n = \sum_v C_n(v)$, so the magnitude of the derivative of the penalty should be less than the magnitude of the derivative of the gain, at least in the limit.}
The penalty term changes mostly as a function of time: each selected sentence increases the size of the selected (cynical) corpus by roughly the average number of words per sentence. 
By contrast, the sentence gain estimate changes as a function of which particular sentences have been chosen. There is no equivalent notion of ``average vocabulary content of a sentence'', so the gain terms continue to vary significantly according to the sentence. 

The converging penalty terms mean that we can get away with sorting and storing sentences by their estimated gain rather than their estimated $\Delta H$ score, and compute the penalty on the fly. There may be a few cases where ordering by $\Delta H$ score would not match the ordering by estimated gain, but on the whole it is good enough. A sentence with a pleasing gain estimate is probably useful, even if it is long enough (or the current $n+1^{th}$ iteration is early enough) to have a large penalty term.

\begin{enumerate}
\item Estimate the $\underset{n\rightarrow n+1}{\textrm{Gain}} (v) \approx \frac{C_{\textsc{repr}}(v)}{W_{\textsc{repr}}} \log \frac{C_n(v)}{C_n(v) + 1}$ for the  $\approx$10,000 word types in $V_{+REPR}$.
\item Sort the word types in $V_{+REPR}$ by their estimated gain, and note the best word $v'$
\item Let $\textsc{Avail}(v')$ be the set of lines still in \textsc{Available} that contain $v'$.\\
The \underline{previous iteration} left this list sorted by estimated sentence gain. We can update the sentence gain score for only the first sentence on the list, and then note where in the list it would be resorted to. Let the formerly-best sentence's new index in the list be $i$. Thanks to the monotonicity of the sentence gain estimate,  only sentences with index $< i$ need to have their scores updated.\\
Compute the $\underset{n\rightarrow n+1}{\Delta H}$ score for each such sentence in $\textsc{Avail}(v')$.\\
\item Sort $\textsc{Avail}(v')$ by $\underset{n\rightarrow n+1}{\Delta H}$ score, and choose the sentence with the best (lowest) score to be sentence $n+1$. Remove it from \textsc{Available} and add it to \textsc{Jaded}.
\end{enumerate}

It does not seem like the complexity cost changes appreciably.

\subsection{Good-Enough Code \nopunct \\}

The process of removing the chosen sentence from \textsc{Available} is not efficient. Recall that we are only looking at lines that contain our best word $v'$. For each of the 10,000 or so words in $V_{+REPR}$, we maintain a list of all sentences that contain it. These lists enable the significant complexity reduction in Section \ref{sec:good-enough-sentences}.  Every sentence thus appears as many times in the full mapping as there are words types in the sentence. 

When we remove the best sentence from \textsc{Available} at each iteration, we are simply using \texttt{shift} to remove the first element from the list of sentences for the single best word. The other mentions of that sentence, in the sentence-lists for the other words in the sentence, are \underline{not} removed. Each sentence-list is kept sorted by the estimated gain of the sentences in it. Therefore there is not a cheap way to find and remove a sentence without checking every sentence in the list.  

Instead, we maintain a small hash table in which the sentence IDs are the keys, and the value is ``true''. When we remove the best sentence from \textsc{Available}, we delete its key from the hash. The entries for that sentence in the other lists are unchanged, but they are now ghosts of sentences already-selected. We eliminate these ghosts only when we happen to see them, namely when we (try to) update the sentence's score or select the same sentence again.

When updating a sentence's score, it is simple enough to check whether the sentence ID is still in the hash table. This operation is O(1), and we do it as many times as there are sentences to check. If we checked every sentence, we would not save any time. In Section~\ref{sec:good-enough-scoring} we mentioned using the change in index after sorting as a way to reduce the number of sentences to check-- here is why this reduction matters. The worst case scenario is that the sentences always get resorted to the end of the list, so the  complexity cost does not change even though in practice it is faster.

\section{Batch Mode = Fast Mode \nopunct \\}

Finding the single next best sentence to add is the most efficient strategy for the case where we really want to pick the absolute best sentences for manual processing (transcription, translation, etc), because the external cost for using each sentence is relatively high.

However, it is common to just want \underline{a} relevant subcorpus for automatically training a language model, an MT system, or some other downstream system. Here the cost for using each sentence is relatively low, and it only matters whether a particular sentence is selected (or not). It does not matter much whether the sentence is chosen on the thirtieth or three millionth step, and certainly not if it was number 235,442 versus number 235,443. As such, there is little point to worrying about getting the best sentence at the optimal iteration. Any iteration will do, much like any port in a storm.

We enable Batch Mode to hurry the process along a bit. Let $A(v') = |\textsc{Avail}(v')|$ as the number of available sentences still containing the best word $v'$. At every iteration, we now always update the scores of the top $\sqrt{A(v')}$ lines (not counting ghosts that we prune). We then always select (and remove) the top $\frac{1}{2}\sqrt{A(v')}$ lines.

This reduces the complexity even further. We previously estimated $A(v') \approx N^{\frac{1}{3}}$, so:

$$O( \frac{1}{\frac{1}{2}\sqrt{A(v')}} N^{\frac{4}{3}} \log N) \approx O( 2 \cdot \frac{N^{\frac{4}{3}}}{N^{\frac{1}{3}}} \log N) \approx O( N \log N) $$

This is about as good as one can hope for.

As string-identical sentences have identical estimated gains and penalties, Batch Mode will often select multiple copies of the same sentence in a single batch (the one-at-a-time version will never do this). Picking a sentence twice (or twenty times) is not a bad thing, as long as the sentence contains at least one word that needs to be seen many more times in order to better match the \textsc{Representative} distribution. As the sentences are selected based on the presence of such a word, the condition is satisfied by construction.

However, it greatly bothers human users to see The Algorithm make such an obvious `mistake', and it leads to bikeshedding. We are tired of explaining how statistical methods work to people who think that a second copy of a sentence invalidates the entire selection method.
Therefore, we unique all sentences within a batch, and un-select any copies. The un-selected identical sentences go back onto the word's list of un-selected sentences (\textit{e.g.} $A(v')$). Thus it is still both possible and probable for 500 occurrences of something as banal as \texttt{``Thank you .''} to be selected. Mercifully, these copies will be sprinkled across 500 batches in the final selected corpus, so fewer humans will notice.

\section{Wrapping Up \\}
All experimental results will be published separately, for entirely amiable legal reasons. The code for cynical data selection is going through proper corporate/legal channels for release. It will be on github with a permissive license as soon as that is completed (expected September 2017). In a pinch, this work contains enough detail to implement the algorithm.

The method presented in the preceding sections has the following properties:

\begin{itemize}
\item Is responsive to the extent to which two corpora differ. \\
It retains the core concept from cross-entropy difference of playing the two distributions against each other, but only for the parts that actually differ between the two distributions.

\item Quickly reaches near-optimal vocabulary coverage.\\
The cynical selection method selects sentences by first looking for word whose probability estimate can be improves the most. 
It happens that the greatest possible improvement is to go from zero -- unseen -- to a bad estimate (seen once). Because of this, it winds up selecting almost every word in the \textsc{Representative} corpus-- exactly once! -- very early in the process. This minimizes the number of unknown words in the training corpus, meaning the cynical subset preserves the breadth of \underline{useful} vocabulary found in the much larger \textsc{Available} corpus.
This is also an improvement over Moore-Lewis: empirically, the cynical subsets have about 80\% fewer OOV (out of vocabulary) tokens than the Moore-Lewis subsets.

\item Takes into account what has already been selected. \\
This is the key benefit of the submodular data selection methods of \cite{KirchhoffBilmes:2014} and \cite{WeiLiuKBB:2014}. Our cynical method seems to satisfy the definition of submodularity\footnote{ from \cite{WeiLiuKBB:2014}: ``...an equivalent definition of submodularity is $f(j|S) \ge f(j|T), \forall S ? T$. That is, the incre- mental gain of adding item $j$ to the set decreases when the set in which $j$ is considered grows from $S$ to $T$." In our case, the gain of adding a sentence decreases as the already-selected set grows.} However, we are able to avoid much of their heavy lifting by always using the same feature function (entropy gain) and not needing to specify a budget. We can automatically pick a good cutoff point by select sentences until the estimated gain becomes (and stays) positive.

\item Does not involve defining any kind of domain. \\
More importantly, it also does not define what it means to \underline{not} belong to the domain.
A sentence could be moved from one corpus to another, or deleted, or cloned into both, without changing the definition of the \textsc{Representative} and \textsc{Unadapted} distributions. Those distributions represent the two corpora, regardless of what they may contain. Compare with the Moore-Lewis method, where the Pool data had to differ significantly from the Task corpus, and subsequent classifier-based extensions where ``not-in-domain" data had to be defined.
Similarly, our method does not involve any kind of classifier. We can quantify relevance without needing to label anything.

\item Has real units. \\
A sentence's relevance score is the change it has, in bits of entropy, on how well a model can represent the \textsc{Representative} data. As they are real units, they can be compared across systems, methods, etc. Cross-entropy difference scores are also in bits, but they are relative: ``by how many more bits does the in-domain LM like the sentence than the Pool LM does?''. Changing either of the models renders the scores meaningless. Other methods use abstract measures like cosine similarity in higher-dimensional spaces, or other normalized scores that lack units entirely. Three may be the number of the count\footnote{ ``...and the number of the counting shall be three."}, but it makes little sense to say ``this sentence is 3 close to the in-domain model".

\item Knows approximately when to stop. \\
Data selection methods often require gridsearch to determine the best amount of data to select. First try 1\%, then 5\%, then 10\% and so on, build systems on each of them, and take the one that works the best. The optimal value does not carry over between systems nor language arcs.
The cynical method knows it has selected enough data for the most good enough (if not best) subset: when $\underset{n\rightarrow n+1}{\Delta H}$ becomes (and stays) positive. No tuning nor searching is needed.
\end{itemize}

\bibliographystyle{apalike}
\bibliography{library}
\end{document}